\documentclass{article}
\usepackage{spconf,amsmath,graphicx}

\usepackage[mathscr]{eucal}

\usepackage{multirow}
\usepackage{makecell}
\usepackage{amssymb}
\usepackage{pifont}
\usepackage{nccmath}
\def\x{{\mathbf x}}

\def\y{{\mathbf y}}

\def\etal{{\em et al.~}}

\makeatother
\usepackage[normalem]{ulem}
\usepackage{xcolor}

\usepackage{hyperref} 

\usepackage[backend=bibtex,citestyle=numeric-comp,bibstyle=ieee,sorting=nty,defernumbers=true,giveninits=true,doi=false,isbn=false,url=false,eprint=false,minbibnames=5,maxbibnames=5]{biblatex}
\usepackage{tabularx}
\usepackage{booktabs}
\usepackage{multirow}
\usepackage{ragged2e}
\usepackage{bm}

\usepackage{boldline}

\newcolumntype{C}{>{\Centering\arraybackslash}X}

\newcolumntype{t}{>{\Centering\hsize=.6\hsize}X}
\newcolumntype{s}{>{\Centering\hsize=.5\hsize}X}
\newcolumntype{j}{>{\Centering\hsize=.4\hsize}X}
\newcolumntype{k}{>{\Centering\hsize=.3\hsize}X}
\newcolumntype{y}{>{\Centering\hsize=.1\hsize}X}
\newcolumntype{z}{>{\Centering\hsize=.1\hsize}X}

\newcolumntype{e}{>{\raggedright\hsize=.35\hsize}X}
\newcolumntype{v}{>{\raggedright\hsize=.55\hsize}X}
\newcolumntype{u}{>{\raggedright\hsize=.7\hsize}X}
\title{End-to-end audio-visual speech recognition with Conformers}

\name{Pingchuan Ma, Stavros Petridis, Maja Pantic
\address{
Department of Computing, Imperial College London, UK
 }}

\begin{document}
\maketitle
\ninept
\setlength{\abovedisplayskip}{2pt}
\setlength{\belowdisplayskip}{2pt}

\begin{abstract}
In this work, we present a hybrid CTC/Attention model based on a  ResNet-18 and Convolution-augmented transformer (Conformer), that can be trained in an end-to-end manner. In particular, the audio and visual encoders learn to extract features directly from raw pixels and audio waveforms, respectively, which are then fed to conformers and then fusion takes place via a Multi-Layer Perceptron (MLP). The model learns to recognise characters using a combination of CTC and an attention mechanism. We show that end-to-end training, instead of using pre-computed visual features which is common in the literature, the use of a conformer, instead of a recurrent network, and the use of a transformer-based language model, significantly improve the performance of our model. We present results on the largest publicly available datasets for sentence-level speech recognition, Lip Reading Sentences 2 (LRS2) and Lip Reading Sentences 3 (LRS3), respectively. The results show that our proposed models raise the state-of-the-art performance by a large margin in audio-only, visual-only, and audio-visual experiments.
\end{abstract}
\begin{keywords}
audio-visual speech recognition, end-to-end training, convolution-augmented transformer
\end{keywords}
 
\section{Introduction}

Audio-Visual Speech Recognition (AVSR) is the task of transcribing text from audio and visual streams, which has recently attracted a lot of research attention due to its robustness against noise. Since the visual stream is not affected by the presence of noise, an audio-visual model can lead to improved performance over an audio-only model as the level of noise increases.

Traditional audio-visual speech recognition methods follow a two-step approach, feature extraction and recognition \cite{Potamianos2003, Dupont2000}. 
Several End-to-End (E2E) approaches have been recently presented by combining feature extraction and recognition inside a deep neural network, and this has led to a significant improvement in Visual Speech Recognition (VSR) and Automatic Speech Recognition (ASR), respectively. In VSR, Assael \etal \cite{assael2016lipnet} developed the first end-to-end network based on 3D convolution with Gated Recurrent Units (GRUs) for recognising visual speech on GRID \cite{cooke2006audio}. Shillingford \etal \cite{shillingford2019large} proposed an improved version of the model called Vision to Phoneme (V2P) which predicts phoneme distributions, instead of characters, from video clips. Chung and Zisserman \cite{chung2017lip} developed an attention-based sequence-to-sequence model for VSR in-the-wild. Zhang \etal \cite{zhang2019spatio} proposed a Temporal Focal block to capture temporal dynamics locally in a convolution-based sequence-to-sequence model. In ASR, \cite{zeghidour2018end, parcollet2020e2e} have been recently shown to achieve better recognition performance by replacing the hand-crafted features such as log-Mel filter-bank features with deep representations from networks.

Several audio-visual approaches have been recently presented where pre-computed visual or audio features are used \cite{afouras2018deep, makino2019recurrent, yu2020audio, petridis2018audio, DBLP:journals/taslp/SterpuSH20}. Afouras \etal developed a transformer-based sequence-to-sequence model by using pre-computed visual features and log-Mel filter-bank features as inputs. \cite{makino2019recurrent, yu2020audio, DBLP:journals/taslp/SterpuSH20} focus on using video clips and log-Mel filter-bank features as inputs to train an audio-visual speech recognition model in an end-to-end manner. Few audiovisual studies are truly E2E, in the sense that they are trained with raw pixels and audio waveforms \cite{petridis2018end, ma2019investigating}. In particular, \cite{petridis2018end} was applied only to word classification while \cite{ma2019investigating} was tested on a constrained environment.

In this work, we extend our previous audio-visual model presented in \cite{petridis2018audio} to an end-to-end model, which extracts features directly from raw pixels and audio waveform, and introduce a few changes which significantly improve the performance. In particular, we integrate the feature extraction stage with the hybrid CTC/attention back-end and train the model jointly. This results in a significant improvement in performance. We also replace the recurrent networks with conformers, which further push the state-of-the-art performance. Finally, we replace the RNN-based Language Model (RNN-LM) with a transformer-based LM which enhances the performance even more. We also perform a comparison between audio-only models trained with log-Mel filter-bank features and raw waveforms.  Although in clean conditions they both perform similarly, the raw audio model performs slightly better in noisy conditions. We evaluate the proposed architecture on the largest in-the-wild audio-visual speech datasets, LRS2 and LRS3. The state-of-the-art performance is raised by a large margin for audio-only, visual-only and audio-visual experiments on both datasets, even outperforming methods trained on much larger external datasets.

\looseness - 1
\section{Datasets}
For the purpose of this study, we use two large-scale publicy available audio-visual datasets, LRS2 \cite{chung2017lip} and LRS3 \cite{afouras2018LRS3}. Both datasets are very challenging as there are large variations in head pose and illumination.
LRS2 \cite{chung2017lip} consists of 224.1 hours with 144\,482 video clips from BBC programs. In particular, there are 96\,318 utterances for pre-training (195\,hours), 45\,839 for training (28\,hours), 1\,082 for validation (0.6\,hours), and 1\,243 for testing (0.5\,hours).

LRS3 \cite{afouras2018LRS3} collected from TED and TEDx talks is twice as large as the LRS2 dataset. LRS3 contains 151\,819 utterances (438.9\,hours). Specifically, there are 118\,516 utterances in the pre-training set (408\,hours), 31\,982 utterances in the training-validation set (30\,hours) and 1\,321 utterances in the test set (0.9\,hours).

\looseness - 1
\section{ARCHITECTURE}
\begin{figure}[!t]
    \centering
    \includegraphics[width=.7\columnwidth]{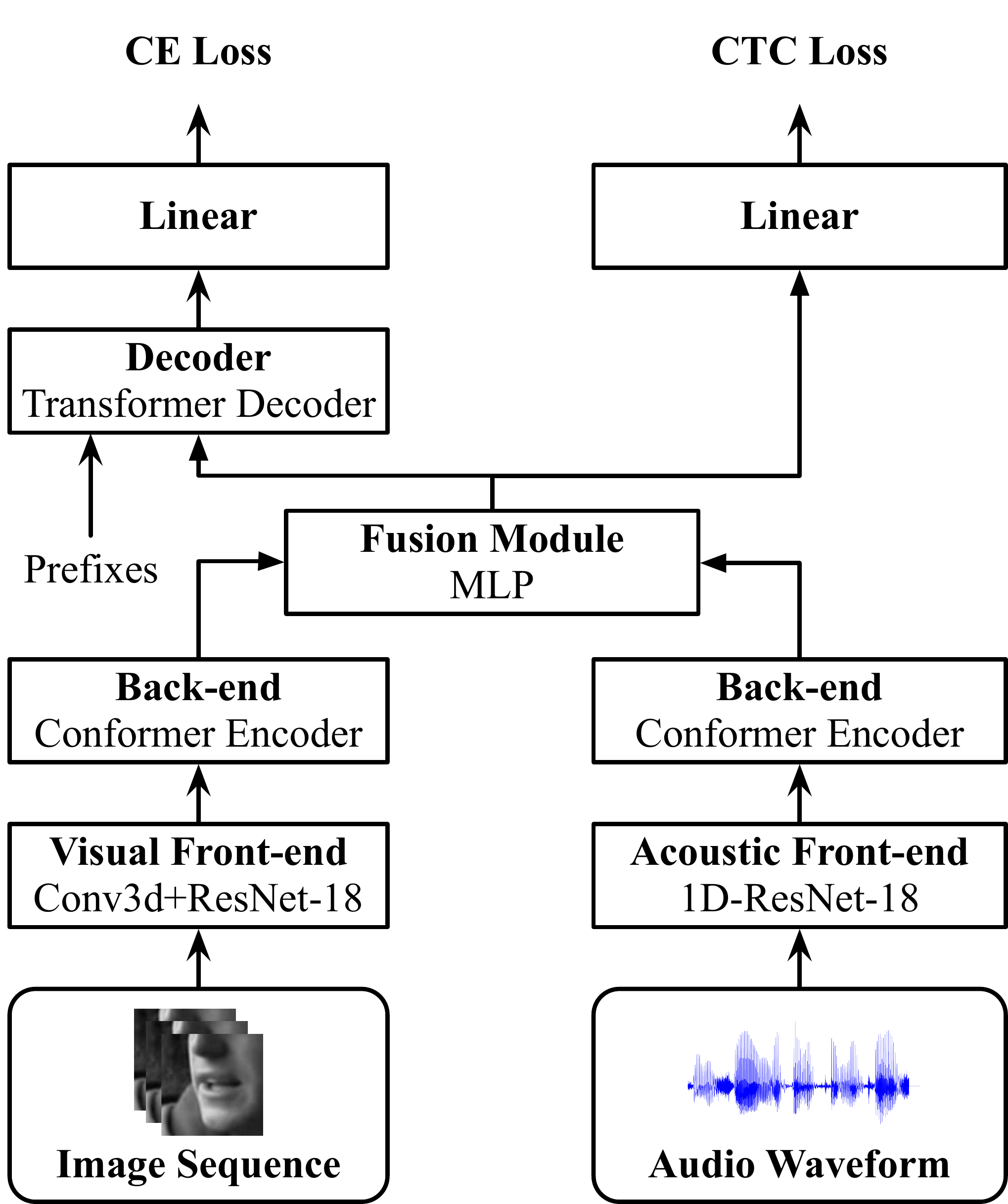}
    \caption{End-to-end audio-visual speech recognition architecture. The inputs are pixels and raw audio waveforms.}
    \label{fig:av_architecture}
    \vspace{-5mm}
\end{figure}

The proposed architecture for audio-visual speech recognition is shown in Fig.~\ref{fig:av_architecture}. The encoder of the audio-visual model is comprised of three components, the front-end, the back-end, and the fusion modules, as explained below.

\noindent\textbf{Front-end}\quad
The acoustic and visual front-ends architectures are shown in Table~\ref{net:resnet}. For the visual stream, we use a modified ResNet-18 \cite{he2016deep, stafylakis2017combining} in which the first convolutional layer is replaced by a 3D convolutional layer with a kernel size $5 \times 7 \times 7$. The visual features at the end of the residual block are squeezed along the spatial dimension by a global average pooling layer. For the audio stream, we use a ResNet-18 based on 1D convolutional layers, where the filter size at the first convolutional layer is set to 80 (5ms). To down-sample the time-scale, the stride is set to 2 at every block. The only exception is the first block, where we set the stride to 4. At the end of the front-end module, acoustic features are down-sampled to 25 frames per second so the match the frame rate of the visual features.

\looseness-1
\noindent\textbf{Back-end}\quad
We use the recently proposed conformer encoder \cite{gulati2020conformer} as the back-end for temporal modeling. It is comprised of an embedding module, followed by a set of conformer blocks. In the embedding module, a linear layer projects the features from ResNet-18 to a $d_k$-dimensional space. The projected features are encoded with relative position information \cite{dai2019transformer}. In each conformer block, a feed-forward module, a Multi-Head Self-Attention (MHSA) module, a convolutional module, and a feed-forward module are stacked in order.
In particular, the feed-forward module is composed of a $d^{{\rm ff}}$-dimensional linear layer, followed by Rectified Linear Units (ReLU), a dropout layer, and a second linear layer with an output size of $d^k$.
The MHSA module receives queries $Q$, keys $K$, and values $V$ as inputs, where $Q\in \mathbb{R}^{T\times d_k}$, $K\in \mathbb{R}^{T\times d_k}$, and $V\in \mathbb{R}^{T\times d_v}$, $T$ denotes the sequence length and $d_k$ and $d_v$ are the dimensions for queries/keys and values, respectively. Suppose $Q=K=V$ in the encoder and $W_i^Q$, $W_i^K$ and $W_i^V$ are denoted as the weights of linear transformation for Q, K and V, respectively, the matrix of outputs at $i$-th head self-attention is computed through Scaled Dot-Product Attention \cite{vaswani2017attention}:
$f_i(Q_i^\prime, K_i^\prime, V_i^\prime)={\rm softmax}(Q_i^\prime K_i^{\prime T})/d_k^{0.5}V_i^\prime$
, where $Q_i^\prime=QW_i^Q, K_i^\prime=KW_i^K, V_i^\prime=VW_i^V$.
\looseness-1
The convolutional module contains a point-wise convolutional layer with an expansion factor of 2, followed by Gated Linear Units (GLU) \cite{dauphin2017language}, a temporal depth-wise convolutional layer, a batch normalisation layer, a swish activation layer, a point-wise convolutional layer, and a layer normalisation layer. This combination has been shown to improve ASR performance compared to the transformer architecture as it better captures temporal information locally and globally \cite{gulati2020conformer}.

\begin{table}[!tb]
\small
\renewcommand\arraystretch{1.4}
\begin{center}{\scalebox{0.83}{
\begin{tabular}{c|p{3.5cm}<{\centering}|p{3.5cm}<{\centering}}
\hlineB{2}
stage&
Input audio waveform\linebreak($ T_a \times 1$) 
&Input image sequence\linebreak($ T_v    \times W \times H$) \\
\hline

\multirow{ 2}{*}{${\rm conv}_1$}&
\multirow{ 2}{*}{${\rm conv1d}, 80, 64, {\rm stride}\; 4$} 
&
\multicolumn{1}{c}{${\rm conv3d}, 5\times 7^2, 64, {\rm stride}\; 1\times2^2$}    \\
\cline{3-3}
&
&
\multicolumn{1}{c}{${\rm maxpool}, 1\times 3^2$}    \\
\cline{1-3}
${\rm res}_2$&
\multicolumn{1}{c|}
{$\begin{bmatrix} {\rm conv1d}, 3, 64 \\ {\rm conv1d}, 3, 64 \\ \end{bmatrix} \times 2$} 
&
{$\begin{bmatrix} {\rm conv2d}, 3^2, 64 \\ {\rm conv2d}, 3^2, 64 \\ \end{bmatrix} \times 2$}    \\
\cline{1-3}

${\rm res}_3$&
\multicolumn{1}{c|}
{$\begin{bmatrix} {\rm conv1d}, 3, 128 \\ {\rm conv1d}, 3, 128 \\ \end{bmatrix} \times 2$} 
& {$\begin{bmatrix} {\rm conv2d}, 3^2, 128 \\ {\rm conv2d}, 3^2, 128 \\ \end{bmatrix} \times 2$}    \\
\cline{1-3}

${\rm res}_4$&
\multicolumn{1}{c|}
{$\begin{bmatrix} {\rm conv1d}, 3, 256 \\ {\rm conv1d}, 3, 256 \\ \end{bmatrix} \times 2$} 
& {$\begin{bmatrix} {\rm conv2d}, 3^2, 256 \\ {\rm conv2d}, 3^2, 256 \\ \end{bmatrix} \times 2$}    \\
\cline{1-3}

${\rm res}_5$&
\multicolumn{1}{c|}
{$\begin{bmatrix} {\rm conv1d}, 3, 512 \\ {\rm conv1d}, 3, 512 \\ \end{bmatrix} \times 2$} 
& {$\begin{bmatrix} {\rm conv2d}, 3^2, 512 \\ {\rm conv2d}, 3^2, 512 \\ \end{bmatrix} \times 2$}    \\
\cline{1-3}

${\rm pool}_6$&
\multicolumn{1}{c|}{${\rm average\;pooling, stride}\; 20$}
& \multicolumn{1}{c}{${\rm global\;average\;pooling}$} \\
\cline{1-3}

\hlineB{2}
\end{tabular}}}
\end{center}
\normalsize
\vspace{-3mm}
\caption{The architecture of acoustic and visual Front-end. The dimensions of kernels are denoted by $\{ {\rm temporal\;size}\times {\rm spatial\;size}^2, {\rm channels} \}$. The acoustic model and visual backbones have 3.85\,M and 11.18\,M parameters, respectively. $T_a$ and $T_v$ denote the number of input samples and frames, respectively.}
\label{net:resnet}
\vspace{-6mm}

\end{table}

\noindent\textbf{Fusion Layers}\quad
The acoustic and visual features from the back-end modules are then concatenated and projected to $d_k$-dimensional space by an MLP. The MLP is composed of a linear layer with an output size of $4 \times d_k$ followed by a batch normalization layer, ReLU, and a final linear layer with an output dimension $d_k$.

\noindent\textbf{Decoder}\quad
We use the transformer decoder proposed in \cite{vaswani2017attention}, which is composed of an embedding module, followed by a set of multi-head self-attention blocks. In the embedding module, a sequence of the prefixes from index 1 to $l-1$ is projected to embedding vectors, where $l$ is the target length index. The absolute positional encoding \cite{vaswani2017attention} is also added to the embedding. A self-attention block is comprised of two attention modules and a feed-forward module. Specifically, the first self-attention module uses $Q=K=V$ as input and future positions at its attention matrix are masked out. The second attention module uses the features from the previous self-attention module as $Q$ and the representations from the encoder as $K$ and $V$ ($K=V$). The component in the feed-forward module is the same as in the encoder.

\noindent\textbf{Loss functions}\quad
Let $\x=[x_1, ..., x_T]$ and $\y=[y_1, ..., y_L]$ be the input sequence and target symbols, respectively, with $T$ and $L$ representing the input and target lengths, respectively. Recent works in audio-visual speech recognition rely mostly on CTC \cite{ma2019investigating} or attention-based models \cite{chung2017lip, afouras2018deep} for audio-visual recognition. CTC loss assumes conditional independence between each output prediction and has a form of $p_{\text{\tiny CTC}}(\y|\x) \approx \prod_{t=1}^Tp(y_t|\x)$. An attention-based model gets rid of this assumption by directly estimating the posterior on the basis of the chain rule, which has a form of $p_{\text{\tiny CE}}(\y|\x) = \prod_{l=1}^Lp(y_l|y_{<l}, \x)$. In this work, we adopt a hybrid CTC/Attention architecture \cite{watanabe2017hybrid} to force monotonic alignments and at the same time get rid of the conditional independence assumption. The objective function is computed as follows:
\begin{align}
\mathcal{L}\!=\!\alpha\text{log}p_{\text{\tiny CTC}}(\y|\x)\!+\!(1\!-\!\alpha)\text{log} p_{\text{\tiny CE}}(\y|\x)
\label{eq:trainingCTCweight}
\end{align}

where $\alpha$ controls the relative weight in CTC and attention mechanisms.

\looseness - 1
\section{Experiments}

\noindent\textbf{Pre-processing}\quad
In each video, 68 facial landmarks are detected and tracked using dlib \cite{king2009dlib}. To remove differences related to rotation and scale, the faces are aligned to a neural reference frame using a similarity transformation. A bounding box of $96 \times 96$ is used to crop the mouth ROIs. The cropped patch is further converted to gray-scale and normalised with respect to the overall mean and variance on the training set. Each raw audio waveform is normalised by removing its mean and dividing by its standard deviation.

\noindent\textbf{Data augmentation}\quad
Following \cite{stafylakis2017combining, martinez2020lipreading}, random cropping with a size of $88 \times 88$ and horizontal flipping with a probability of 0.5 are performed for each image sequence.
For each audio waveform, additive noise, time masking, and band
reject filtering are performed in the time domain. Babble noise from the NOISEX corpus \cite{varga1993assessment} is added to the original audio clip with an SNR level from [-5\,dB, 0\,dB, 5\,dB, 10\,dB, 15\,dB, 20\,dB]. The selection of one of the noise levels or the use of a clean waveform is done using a uniform distribution. Similarly to \cite{, kharitonov2020data}, 2 sets of consecutive audio samples with a maximum length of 0.4 seconds are set to zero and 2 sets of consecutive frequency bands with a maximum width of 150 Hz are rejected. In audio-only experiments, we add speed perturbation by setting the speed between 0.9 and 1.1.

\noindent\textbf{Experimental settings}\quad
The network is initialised randomly, with the exception of the front-end modules in the encoder part, which in some experiments are initialised based on the publicly available pre-trained models on LRW \cite{DBLP:journals/corr/abs-2007-06504} \footnote{Pre-trained LRW models are available at \url{https://sites.google.com/view/audiovisual-speech-recognition}}.
The back-end modules use a set of hyper-parameters ($e\!=\!12$, $d^{{\rm ff}}\!=\!2048$, $d^{{\rm k}}\!=\!256$, $d^{{\rm v}}\!=\!256$), where $e$ denotes the number of conformer blocks. The number of heads $n^{{\rm head}}$ is set to 4 in visual-only models and 8 in audio-only/audio-visual models, respectively. Kernel size is set to 31 in each depth-wise convolutional layer. The transformer decoder uses 6 self-attention blocks, where the hyper-parameters settings in feed-forward and self-attention modules are the same as in the encoder. The Adam optimizer \cite{kingma2014adam} with $\beta_{1}\!=\!0.9$, $\beta_{2}\!=\!0.98$ and $\epsilon\!=\!10^{-9}$ is used for end-to-end training with a mini-batch size of 8.
Following \cite{vaswani2017attention}, the learning rate increases linearly with the first 25\,000 steps, yielding a peak learning rate of 0.0004 and thereafter decreases proportionally to the inverse square root of the step number. The whole network is trained for 50 epochs. Note that the utterances with more than 600 frames in the pre-training set are excluded during training.

\noindent\textbf{Language Model}\quad
We train a transformer-based language model \cite{irie2019language} for 10 epochs. The language model is trained by combining the training transcriptions of LibriSpeech (960\,h) \cite{panayotov2015librispeech}, pre-training and training sets of LRS2 \cite{chung2017lip} and LRS3 \cite{afouras2018LRS3}, with a total of 16.2 million words. The weighted prior score from the language model is incorporated through a shallow fusion, which is described in Eq.~\ref{eq:decode}.
\begin{align}
\nonumber
\hat{\y}\!=\!\mathop\mathrm{argmax} \limits_{\y \in \hat{\mathcal{Y}}} \,\{\!\lambda\text{log}p_{\text{\tiny CTC}}(\y|\x)\!&+\!(1\!-\!\lambda)\text{log}p_{\text{\tiny CE}}(\y|\x)\!\\[-8pt]
&+\!\beta\text{log}p_{\text{\tiny LM}}(\y)\!\}
\label{eq:decode}
\end{align}
where $\hat{\mathcal{Y}}$ is a set of predictions of target symbols. $\lambda$ is a relative CTC weight at the decoding phase, and $\beta$ is the relative weight for the language model. In our work, we set $\lambda$ to 0.1 and $\beta$ to 0.6, respectively.

\looseness - 1
\section{Results}
\noindent\textbf{Ablation Studies}\quad
\begin{table}[!t]
\centering
\begin{tabularx}{.9\columnwidth}{u k}
\toprule
Method & WER \\
\midrule
Baseline \cite{petridis2018end}  &63.5 \\
\enskip\enskip + E2E &50.9 \\
\enskip\enskip\enskip\enskip + LRW pre-training & 46.2 \\
\enskip\enskip\enskip\enskip\enskip\enskip + Conformer encoder &42.4 \\
\enskip\enskip\enskip\enskip\enskip\enskip\enskip\enskip + Transformer LM & 37.9 \\
\bottomrule
\end{tabularx}
\caption{Ablation study on visual speech recognition performance on LRS2.}
\label{ablation_study_on_LRS2}
\vspace{-1mm}
\end{table}
In this section, we investigate the impact of each change on the baseline hybrid CTC/Attention model \cite{petridis2018audio}. Results on LRS2 are shown in Table~\ref{ablation_study_on_LRS2}.
We first train a model from scratch in an end-to-end manner, resulting in an absolute improvement of 12.6\,\% over the two stage approach, where visual features are first extracted and then fed to the back-end. We initialise the visual front-end with a model pre-trained on LRW and a further absolute improvement of 4.7\,\% is observed. Then, we replace the LSTM encoders and decoders with a conformer encoder and a transformer decoder, respectively, which results in an absolute improvement of 3.8\,\%. We also replace the RNN-based language model with a transformer-based language model and achieve a WER of 37.9\,\%. This leads to an absolute improvement of 4.5\,\%.

\begin{figure}[t]
    \centering
    \includegraphics[width=.95\columnwidth]{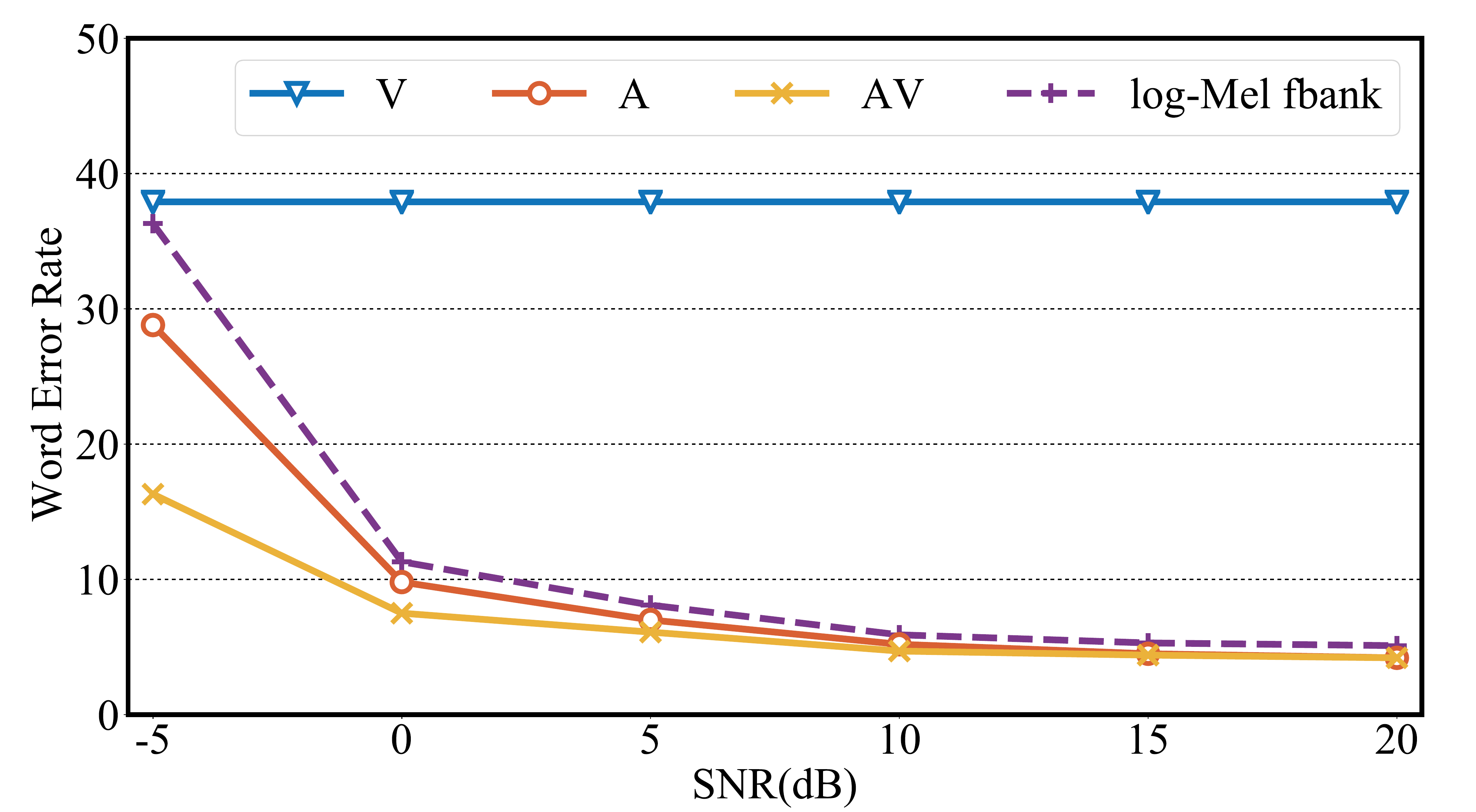}
\vspace{-2mm}
\caption{Word Error Rate (WER) as a function of the noise
level. A: End-to-End audio model. V: End-to-End visual
model, AV: End-to-End audio-visual model. log-Mel filter-bank: A conformer model trained with log-Mel filter-bank features.}
\label{fig:av_results}
\vspace{-3mm}
\end{figure}

\noindent\textbf{Results on LRS2}\quad
Results on LRS2 are reported in Table~\ref{table: results_on_LRS2}. The proposed visual-only model reduces the WER from 48.3\,\% to 39.1\,\%, while using $6$ $\times$ fewer training data \cite{afouras2018deep}. In case, we use the pre-trained LRW model for initialisation the WER drops further to 37.9\,\%. The E2E audio-only model using audio waveforms for training achieves a WER of 4.3\,\%, resulting in an absolute improvement of 2.4\,\%. over the current state-of-the-art. For comparison purposes, we also run an experiment using 80-dimension log-Mel filter-bank features following \cite{petridis2018audio, watanabe2017hybrid}. Similarly to the WavAugment \cite{kharitonov2020data}, we augment the log-Mel filter-bank features via SpecAugment \cite{park2019specaugment}. By replacing the raw audio features with the log-Mel filter-bank features, we observe the same performance, WER 4.3\,\%, which indicates deep acoustic speech representations based on the proposed temporal network can be directly learnt from audio waveforms. To better investigate their differences, we conduct noisy experiments varying different levels of babble noise. The results are shown in Fig.~\ref{fig:av_results}. It is interesting to observe that the performance of the raw audio model slightly outperforms the log-Mel filter-bank based over varying levels of babble noise with a maximum absolute margin of 7.5\,\% 
at -5\,dB. This indicates deep speech representations are more robust to  noise than the log-Mel filter-bank features. In case, we initialise the audio encoder with a model pretrained on LRW then the WER drops to 3.9\,\%.

It is evident that the audio-visual model which directly learns from audio waveforms and raw pixels leads to a small improvement over the audio-only models. We also run audio-only, visual-only, and audio-visual experiments varying the SNR levels of babble noise. The results are shown in Fig~\ref{fig:av_results}. Note that both audio-only and audio-visual models are augmented with noise injection. It is clear that the audio-visual model achieves better performance than the audio-only model. The gap between raw audio-only and audio-visual models becomes larger by the presence of high level of noise. This demonstrates that the audio-visual model is particularly beneficial when the audio modality is heavily corrupted by background noise.

\begin{table}[!t]
\small
\renewcommand\arraystretch{0.9}
\begin{tabularx}{\columnwidth}{e v y}
\toprule
Method &Training Data \,(Hours) &WER \\
\midrule\midrule
\textit{Visual-only} \,($\downarrow$) & &\\
\midrule
MV-WAS \cite{chung2017lip} &LRS2\,(224) &70.4 \\\midrule
LIBS   \cite{zhao2020} &MVLRS\,(730)\,+\,LRS2\,(224)  &65.3 \\\midrule
CTC/Attention \cite{petridis2018audio} &LRW\,(157)\,+\,LRS2\,(224)  &63.5 \\\midrule
Conv-seq2seq  \cite{zhang2019spatio} &LRW\,(157)\,+\,LRS2\&3$^{\text{v0.0}}$\,(698) &51.7  \\\midrule
KD\,+\,CTC  \cite{afouras2020asr} &VC2$^{\text{clean}}$\,(334)\,+\,LRS2\&3$^{\text{v0.4}}$\,(632)  &51.3 \\\midrule
TDNN  \cite{yu2020audio}  &LRS2\,(224)  &48.9 \\\midrule
TM-seq2seq \cite{afouras2018deep}  &MVLRS\,(730)\,+\,LRS2\&3$^{\text{v0.4}}$\,(632)  &48.3  \\\midrule

Ours\,(V) &LRS2\,(224) &\textbf{39.1}\\\midrule
Ours\,(V) &LRW\,(157)\,+\,LRS2\,(224)  &\textbf{37.9} \\
\midrule\midrule
\textit{Audio-only} \,($\downarrow$) & &\\
\midrule
TM-seq2seq \cite{afouras2018deep}  &MVLRS\,(730)\,+\,LRS2\&3$^{\text{v0.4}}$\,(632)  &9.7  \\\midrule
CTC/Attention \cite{petridis2018audio} &LRS2\,(224)  &8.3 \\\midrule
CTC/Attention  \cite{li2020listen} &LibriSpeech\,(960)\,+\,LRS2\,(224)  &8.2 \\\midrule
TDNN  \cite{yu2020audio}  &LRS2\,(224)  &6.7 \\\midrule
Ours\,(filter-bank) &LRS2\,(224)  &\textbf{4.3} \\\midrule
Ours\,(raw A) &LRS2\,(224) &\textbf{4.3} \\ \midrule
Ours\,(raw A) &LRW\,(157)\,+\,LRS2\,(224)  &\textbf{3.9} \\
\midrule\midrule
\textit{Audio-visual} \,($\downarrow$) & & \\
\midrule
TM-seq2seq \cite{afouras2018deep} &MVLRS\,(730)\,+\,LRS2\&3$^{\text{v0.4}}$\,(632)  &8.5  \\\midrule
CTC/Attention \cite{petridis2018audio} &LRW\,(157)\,+\,LRS2\,(224)  &7.0 \\\midrule
TDNN  \cite{yu2020audio}  &LRS2\,(224)  &5.9 \\\midrule
Ours\,(raw A + V) &LRS2\,(224)  &\textbf{4.2}\\\midrule
Ours\,(raw A + V) &LRW\,(157)\,+\,LRS2\,(224)  &\textbf{3.7}\\
\bottomrule
\end{tabularx}

\caption{Word Error Rate \,(WER) of the audio-only, visual-only and audio-visual models on LRS2. VC2$^{\text{clean}}$ denotes the filtered version of VoxCeleb2. LRS2\&3 consists of LRS2 and LRS3. LRS3$^{\text{v0.4}}$ is the updated version of LRS3 with speaker-independent settings.}
\label{table: results_on_LRS2}
\vspace{-4mm}
\end{table}

\begin{table}[!t]
\small
\renewcommand\arraystretch{0.9}
\begin{tabularx}{\columnwidth}{e v y}
\toprule
Method & Training Data \,(Hours) & WER \\
\midrule\midrule
\textit{Visual-only} \,($\downarrow$) & &\\
\midrule
Conv-seq2seq  \cite{zhang2019spatio} &LRW\,(157)\,+\,LRS2\&3$^{\text{v0.0}}$\,(698)  &60.1  \\\midrule
KD\,+\,CTC  \cite{afouras2020asr} &VC2$^{\text{clean}}$\,(334)\,+\,LRS3$^{{\rm v}0.4}$\,(438)  &59.8\\\midrule
TM-seq2seq  \cite{afouras2018deep} &MVLRS\,(730)\,+\,LRS2\&3$^{\text{v0.4}}$\,(632) &58.9  \\\midrule
EG-seq2seq  \cite{xu2020discriminative} &LRW\,(157)\,+\,LRS3$^{\text{v0.0}}$\,(474)  &57.8 \\\midrule
V2P  \cite{shillingford2019large} &YT\,(3\,886)  &55.1 \\\midrule
RNN-T  \cite{makino2019recurrent} &YT\,(31\,000) &33.6 \\\midrule
Ours\,(V) &LRS3$^{\text{v0.4}}$\,(438) &\textbf{46.9} \\\midrule
Ours\,(V) &LRW\,(157)\,+\,LRS3$^{\text{v0.4}}$\,(438) &\textbf{43.3} \\\midrule
Ours\,(V)  &LRW\,(157)\,+\,LRS3$^{\text{v0.0}}$\,(474)  &\textbf{30.4} \\
\midrule\midrule
\textit{Audio-only} \,($\downarrow$) & &\\
\midrule
TM-seq2seq  \cite{afouras2018deep} &MVLRS\,(730)\,+\,LRS2\&3$^{\text{v0.4}}$\,(632) &8.3  \\\midrule
EG-seq2seq  \cite{xu2020discriminative} &LRS3$^{\text{v0.0}}$\,(474)  &7.2 \\\midrule
RNN-T  \cite{makino2019recurrent} &YT\,(31\,000)  &4.8 \\\midrule
Ours\,(filter-bank) &LRS3$^{\text{v0.4}}$\,(438) &\textbf{2.3} \\\midrule
Ours\,(raw A) &LRS3$^{\text{v0.4}}$\,(438) &\textbf{2.3} \\\midrule
Ours\,(raw A) &LRW\,(157)\,+LRS3$^{\text{v0.4}}$\,(438) &\textbf{2.3} \\ \midrule
Ours\,(raw A) &LRW\,(157)\,+LRS3$^{\text{v0.0}}$\,(474)  &\textbf{1.3} \\
\midrule\midrule
\textit{Audio-visual} \,($\downarrow$) & &\\
\midrule
TM-seq2seq  \cite{afouras2018deep} &MVLRS\,(730)\,+\,LRS2\&3$^{\text{v0.4}}$\,(632)  &7.2  \\\midrule
EG-seq2seq  \cite{xu2020discriminative} &LRW\,(157)\,+\,LRS3$^{\text{v0.0}}$\,(474)  &6.8 \\\midrule
RNN-T  \cite{makino2019recurrent} &YT\,(31\,000)  &4.5 \\\midrule
Ours\,(raw A + V) &LRW\,(157)\,+\,LRS3$^{\text{v0.4}}$\,(438)  &\textbf{2.3}\\\midrule
Ours\,(raw A + V) &LRW\,(157)\,+\,LRS3$^{\text{v0.0}}$\,(474)  &\textbf{1.2}\\
\bottomrule
\end{tabularx}
\caption{Word Error Rate \,(WER) of the audio-only, visual-only and audio-visual models on LRS3. VC2$^{\text{clean}}$ denotes the filtered version of VoxCeleb2. LRS2\&3 consists of LRS2 and LRS3. LRS3$^{\text{v0.4}}$ is the updated version of LRS3 with speaker-independent settings.}
\label{table: results_on_LRS3}
\vspace{-4mm}
\end{table}

\looseness - 1
\noindent\textbf{Results on LRS3}\quad
Results on LRS$3^{{\rm v}0.4}$ are reported in Table~\ref{table: results_on_LRS3}.  The best visual-only model has a WER of 43.3\,\%. We observe that our visual-only model outperforms other methods by a large margin while using fewer training data. For the audio-only and audio-visual experiments, our model pushes the state-of-the-art performance to 2.3\,\% and 2.3\,\%, respectively, outperforming \cite{makino2019recurrent} by 2.5\,\% and 2.2\,\%, respectively. It is worth pointing out that our model is trained on a dataset which is 52$\times$ smaller than \cite{makino2019recurrent}, 595 vs 31000 hours.

\looseness-1
We should note that some works use the old version of LRS3 (denoted as v$0.0$), where some speakers appear both in the training and test sets. For fair comparisons, we also report the performance of audio-only, visual-only, and audio-visual model on this version of LRS3 as well. Specifically, the audio-only model achieves a WER to 1.3\,\%. The visual-only model reduces the WER to 30.4\,\%. The audio-visual model reduces the WER to 1.2\,\% which is the new state-of-the-art performance for this set. These significant improvements over LRS3$^{{\rm v}0.4}$ are mainly due to the fact that in LRS3$^{{\rm v}0.0}$ overlapped identities appear in both pre-training and test sets.

\looseness - 1
\section{Conclusions}
In this work, we present an encoder-decoder attention-based architecture for audio-visual speech recognition, which can be trained in an end-to-end fashion and leads to state-of-the-art results on LRS2 and LRS3. Additionally, the audio-visual experiments show that the audio-visual model significantly outperforms the audio-only model especially at high levels of noise. It would also be interesting to investigate in future work an adaptive fusion mechanism that learns to weigh each modality based on the noise levels.

\noindent{\textbf{Acknowledgements.}}\quad We would like to thank Dr. Jie Shen for his help with face tracking. The work of Pingchuan Ma has been partially supported by Honda and ``AWS Cloud Credits for Research''.

\clearpage
\AtNextBibliography{\small}
\section{References}
\begingroup
\setlength\bibitemsep{1pt}
\printbibliography[heading=none]
\endgroup

\end{document}